\documentclass[letterpaper]{article} % DO NOT CHANGE THIS
\usepackage{aaai24}  % DO NOT CHANGE THIS
\usepackage{times}  % DO NOT CHANGE THIS
\usepackage{helvet}  % DO NOT CHANGE THIS
\usepackage{courier}  % DO NOT CHANGE THIS
\usepackage[hyphens]{url}  % DO NOT CHANGE THIS
\usepackage{graphicx} % DO NOT CHANGE THIS
\usepackage{natbib}  % DO NOT CHANGE THIS AND DO NOT ADD ANY OPTIONS TO IT
\usepackage{caption} % DO NOT CHANGE THIS AND DO NOT ADD ANY OPTIONS TO IT
\usepackage{array,booktabs,siunitx}
\usepackage{tabularx}

\usepackage{algorithm}
\usepackage{algorithmic}

\usepackage{graphicx}
\usepackage{multirow}
\usepackage{multicol}
\usepackage{amsmath,amsfonts,bm}
\usepackage{enumitem}
\usepackage{amssymb}
\usepackage{arydshln} % Include the package 
\usepackage{colortbl}

\usepackage{xcolor}
\definecolor{myblue}{RGB}{218,232,252}

\usepackage{newfloat}
\usepackage{listings}
\DeclareCaptionStyle{ruled}{labelfont=normalfont,labelsep=colon,strut=off} % DO NOT CHANGE THIS
\floatstyle{ruled}
\newfloat{listing}{tb}{lst}{}
\floatname{listing}{Listing}
\title{TT-MPD: Test Time Model Pruning and Distillation}
\author{
    Haihang Wu, Wei Wang, Tamasha Malepathirana, Sachith Seneviratne, \\Denny Oetomo, Saman Halgamuge
}
\affiliations{
    The University of Melbourne\\
}

\usepackage{bibentry}
\begin{document}

\maketitle

\begin{abstract}
Pruning can be an effective method of compressing large pre-trained models for inference speed acceleration. Previous pruning approaches rely on access to the original training dataset for both pruning and subsequent fine-tuning. However, access to the training data can be limited due to concerns such as data privacy and commercial confidentiality. Furthermore, with covariate shift (disparities between test and training data distributions), pruning and finetuning with training datasets can hinder the generalization of the pruned model to test data. To address these issues, pruning and finetuning the model with test time samples becomes essential. However, test-time model pruning and fine-tuning incur additional computation costs and slow down the model's prediction speed, thus posing efficiency issues. Existing pruning methods are not efficient enough for test time model pruning setting, since finetuning the pruned model is needed to evaluate the importance of removable components. To address this, we propose two variables to approximate the fine-tuned accuracy. We then introduce an efficient pruning method that considers the approximated finetuned accuracy and potential inference latency saving. To enhance fine-tuning efficiency, we propose an efficient knowledge distillation method that only needs to generate pseudo labels for a small set of finetuning samples one time, thereby reducing the expensive pseudo-label generation cost. Experimental results demonstrate that our method achieves a comparable or superior tradeoff between test accuracy and inference latency, with a 32\% relative reduction in pruning and finetuning time compared to the best existing method.
\end{abstract}

\section{Introduction}
\label{sec:Introduction}

Deep neural networks (DNNs) have demonstrated remarkable performance in computer vision tasks~\cite{He2016DeepRecognition,Lin2020FocalDetection}. However, their substantial computational requirements present challenges for resource-limited devices such as robots and mobile phones~\cite{Liu2022A2020s,Dosovitskiy2021AnScale}. To address this issue, multiple model compression techniques have been proposed to reduce both model size and computational costs~\cite{Tang2022PatchTransformers,Dettmers2022LLM.int8:Scale,Beyer2022KnowledgeConsistent}.

Neural network pruning is a common model compression technique that removes less important elements from a pre-trained network~\cite{Ma2023LLM-Pruner:Models,Zhang2017AnCondensation}. Among model pruning techniques, structured pruning~\cite{Gao2021NetworkMaximization,Yu2023X-Pruner:Transformers}, such as block pruning~\cite{Wang2023PracticalSets}, is particularly attractive as it removes entire contiguous components, enabling real inference time savings. Despite their effectiveness, existing pruning methodologies typically require access to the original training dataset for both pruning and fine-tuning to achieve high accuracy~\cite{Luo2017ThiNet:Compression,Liu2021GroupCompression,Fang2023DepGraph:Pruning,Han2016DeepCoding}.

Two challenges arise when the pre-trained model is pruned and finetuned with the original training dataset. The first challenge arises from limited access to the original training dataset, a common occurrence in scenarios where concerns regarding data privacy (e.g., medical images) and commercial confidentiality (e.g., hidden training details of high-performance models) are prominent. Although few/zero-shot pruning approaches~\cite{Wang2023PracticalSets,Wang2022CompressingReplacing,Kim2020NeuronNeurons} have been developed to address this challenge, their performances are not satisfactory (refer to Table~\ref{exp: few/zero shot pruning}). The second challenge is covariate shift, wherein the distribution of test samples differs from that of training samples. In such cases, pruning decisions based on the training dataset misalign with those derived from the test dataset (Figure~\ref{fig: block importance rank}), and models fine-tuned on the training data  can perform suboptimally on test-time data, leading to unsatisfactory test accuracy (Table~\ref{exp: main results/test time finetuning}). While test time adaptation techniques such as TENT~\cite{Wang2021Tent:Minimization} and EATA\cite{Niu2022EfficientForgetting} can improve test accuracy under covariate shift,  they also increase computational costs, and can reduce inference speed by around 50\% (LS columns in Table~\ref{exp: main results/test time finetuning}). These challenges highlight the importance of using test samples for pruning and fine-tuning.

However, test time model pruning and finetuning incur additional computation costs and slow down the model's prediction speed, posing the efficiency issue. This slowdown is undesirable in scenarios where fast online predictions (e.g., pedestrian identification) are necessary for the safe operation of machines such as self-driving cars.

To improve pruning efficiency, we introduce a test-time pruning method that can remove blocks efficiently and effectively from the pre-trained model. Existing methods~\cite{Luo2020NeuralLimited-Data,Yu2018LearningNetworks,Wang2023PracticalSets} require fine-tuning the pruned model and use the fine-tuned accuracy to assess the impact of removed blocks. However, this process typically numerous forward and backward passes and significantly reduces inference speed. To address this, we identify two computationally inexpensive proxy variables to approximate the fine-tuned accuracy: 1) pruning-induced noise and 2) the model capacity gap between the original and pruned models. Leveraging these variables, we propose a pruning method (Equation~\ref{eq:prune metric}) that can evaluate and prune blocks efficiently.

To enhance finetuning efficiency, we propose an efficient knowledge distillation method to fine-tune the pruned model. Existing knowledge distillation methods use a large teacher model (pre-trained model) to generate pseudo labels, which are then used to fine-tune a smaller student model (the pruned model)~\cite{Hinton2015DistillingNetwork,Wang2022CompressingReplacing}. However, generating pseudo-labels for a large number of samples using the teacher model is computationally expensive. Our findings indicate that a small set of fine-tuning samples with high-dimensional pseudo-labels is sufficient to fine-tune the student model without significant overfitting risks. Building on this insight, we employ the teacher model to generate high-dimensional pseudo-labels for a small fine-tuning dataset and fine-tune the student model accordingly. This approach significantly reduces pseudo-label generation costs, thereby enhancing fine-tuning efficiency.

\begin{figure}[t]
    \centering
        \centering        \includegraphics[width=1.0\linewidth,height=1.5in]{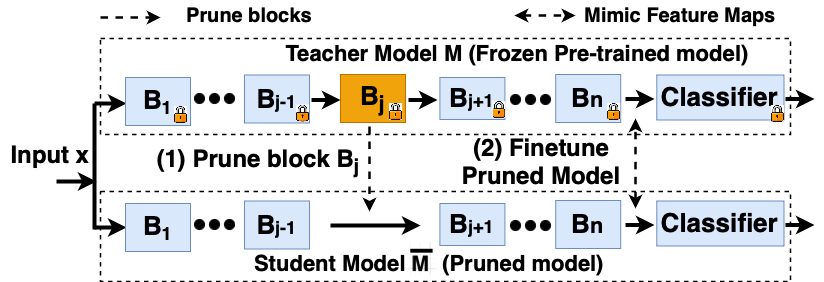}
        \caption{The test-time model pruning and fine-tuning framework consists of two steps: (1) Pruning (dropping) blocks, such as $B_j$, from the pre-trained model $M$  using the proposed pruning criterion (Eq.~\ref{eq:prune metric}), leading to the pruned model $\overline{M}$. (2)  Fine-tuning $\overline{M}$ through knowledge distillation by mimicking the high-resolution feature maps (pseudo labels) from block $B_n$ in the teacher model. After fine-tuning, $M$ is discarded, and $\overline{M}$ is used for inference.}
        \label{fig: overview}
\end{figure}

In summary, our contributions are:

\begin{enumerate}
\item  We propose the first instance of test-time pruning and fine-tuning (Figure~\ref{fig: overview}  and Algorithm~\ref{alg: overview}), demonstrating that using test data for these processes yields superior performance compared to methods using training data in scenarios where access to training data is limited or its distribution differs from that of the test data.
\vspace{-0.1cm}
\item We introduce an efficient test-time pruning method by approximating fine-tuned accuracy using the proposed proxy variables.
\vspace{-0.1cm}
\item We find that high-dimensional pseudo labels enable fine-tuning of the pruned model with a small dataset. Building on this, we introduce an efficient fine-tuning method that generates and stores a small set of high-dimensional pseudo labels, reducing generation costs and enhancing the fine-tuning efficiency of the pruned model.
\vspace{-0.1cm}
\item  Compared to existing approaches, our approach achieves an average 32\% reduction in pruning and fine-tuning time, while offering a comparable or superior balance between test accuracy and inference latency.
\end{enumerate}

\section{Related Works}
\label{sec:related work}

\textbf{Block Pruning} involves the removal of blocks, such as residual blocks, from a model. It demonstrates a superior tradeoff between accuracy and inference latency compared to other structured pruning approaches, such as filter pruning~\cite{Wang2023PracticalSets}. Existing methods~\cite{Luo2020NeuralLimited-Data,Yu2018LearningNetworks,Wang2023PracticalSets} for determining which blocks to prune fall into two categories. The first category makes pruning decisions without fine-tuning the pruned model. They prune the block with minimal impact on the model output. Examples include CURL~\cite{Luo2020NeuralLimited-Data}, which removes the block with the least impact on prediction probability, and $l^2$ ratio~\cite{Yu2018LearningNetworks}, which prunes the block exhibiting the highest similarity between its input and output. The second category removes the block with the least impact on the fine-tuned accuracy~\cite{Wang2023PracticalSets}. Although methods in the second category may achieve higher fine-tuned accuracy compared to those in the first category~\cite{Wang2023PracticalSets}, they are inefficient as the model needs to undergo extensive fine-tuning iterations before pruning decisions can be made. In this study, we aim to enhance the efficiency of approaches in the second category by approximating the finetuned accuracy.

\textbf{Efficient knowledge Distillation:} Knowledge distillation (KD) is a model compression technique that transfers knowledge from a large teacher network to a small student network~\cite{Hinton2015DistillingNetwork,Wang2022CompressingReplacing}. A key efficiency bottleneck in KD is the generation of pseudo-labels by the large teacher network. To improve distillation efficiency, existing works typically filter the fine-tuning samples and query the teacher model for pseudo-labels only for samples that the student network finds uncertain~\cite{Xu2023Computation-EfficientMixup,Wang2020NeuralModel}. However, the student model may overfit these uncertain samples, resulting in low fine-tuned accuracy. In this study, we explore how the student model can be effectively fine-tuned on a small dataset without significant overfitting.

\textbf{Combining Pruning and Knowledge Distillation:} Several works have focused on pruning and distilling large teacher models into smaller student sub-networks~\cite{Sanh2019DistilBERTLighter,Lagunas2021BlockTransformers,Hou2020DynaBERT:Depth,Xia2022StructuredModels,Xu2021RethinkingParadigm,Liang2023HomoDistil:Transformers}.  Some approaches first prune the large model to create a sub-network and then fine-tune this sub-network via knowledge distillation~\cite{Sanh2019DistilBERTLighter,Lagunas2021BlockTransformers,Hou2020DynaBERT:Depth} while others integrate pruning and distillation into a single process~\cite{Xia2022StructuredModels,Xu2021RethinkingParadigm,Liang2023HomoDistil:Transformers}. While these approaches mainly focus on balancing prediction accuracy with inference latency, our work additionally addresses the efficiency challenges of pruning and distillation at test time.

\textbf{Test Time Adaptation:} Test time adaptation refers to the process of adjusting a model, pre-trained on a source domain, to improve prediction accuracy in the test-time domain~\cite{Wang2021Tent:Minimization,Wang2022ContinualAdaptation,Niu2022EfficientForgetting}, but this often leads to reduced inference speed (LS columns in Table~\ref{exp: main results/test time finetuning}). In contrast, our approach focuses on compressing the model at test time to enhance inference speed.

% \textbf{Few Shot Pruning:}

\section{Methods}

\subsection{Problem Definition}

Given a pre-trained neural network $M$ consisting of n blocks, denoted as $\{B_1, \dots, B_n\}$, we remove $N_p$ blocks from $\{B_1, \dots, B_n\}$  and fine-tune the pruned model $\overline{M}$ using test time samples, resulting in the fine-tuned model $\overline{M}_{FT}$. The objectives are:

\begin{itemize}[leftmargin=*,nosep]
  \item  $\min (\frac{\text{Error} (\overline{M}_{FT})}{\text{Inference Latency Saving}})$ (optimize the tradeoff between the test error of $\overline{M}_{FT}$ and the inference latency time saving from pruning)
   \item $\min (T_{prune} + T_{finetune})$ (minimize total time cost for pruning and fine-tuning)
\end{itemize}

\subsection{Efficient Test Time Pruning}
\label{method:Pruning}

\noindent \textbf{Motivation:} The pruning algorithm seeks to minimize both  $\frac{\text{Error} (\overline{M}_{FT})}{\text{Inference Latency Saving}}$ and the pruning time $T_{prune}$ when $N_p$ blocks are removed. While the inference latency time savings from block removal can be calculated offline, the finetuned error $\text{Error} (\overline{M}_{FT})$ requires model fine-tuning, which incurs significant computational costs and increases $T_{prune}$. 

To minimize $T_{prune}$, we identify two computationally inexpensive proxy variables to approximate the fine-tuned error. 
Pruning the pre-trained model introduces initial noise $\epsilon_{ini}$ to the classifier, leading to increased errors. The fine-tuning process, or knowledge distillation, reduces this noise and mitigates the accuracy loss from pruning. Consequently, fine-tuned error can be approximated by the initial noise $\epsilon_{ini}$ and the effectiveness of the finetuning process. 

\textbf{Initial Noise $\epsilon_{ini}$:} When block $B_j$  is pruned, it changes the feature maps input to the classifier. This feature map alteration introduces initial noise $\epsilon_{ini,j}$ to the classifier upon pruning blocks, leading to a decrease in the accuracy~\cite{Bartoldson2020ThePruning}. Consequently, after removing block $B_j$, the initial noise $\epsilon_{ini,j}$ can be measured by the change in block $B_n$'s output feature map from $F_{n}$ (model $M$) to $\overline{F}_{n,j}$ (model $\overline{M}_j$) as shown in Eq.~\ref{eq:output noise}:

\begin{equation} \label{eq:output noise}
\begin{split}
\epsilon_{ini,j} &= MSE (F_{n},\overline{F}_{n,j}) \\
&= \frac{\sum_{i=1}^{P_n} (F_{n,i}-\overline{F}_{n,i,j})^2}{P_n}
\end{split}
\end{equation}

\begin{figure*}[t]
\begin{center}
   \includegraphics[width=1.0\linewidth,height=2.0in]{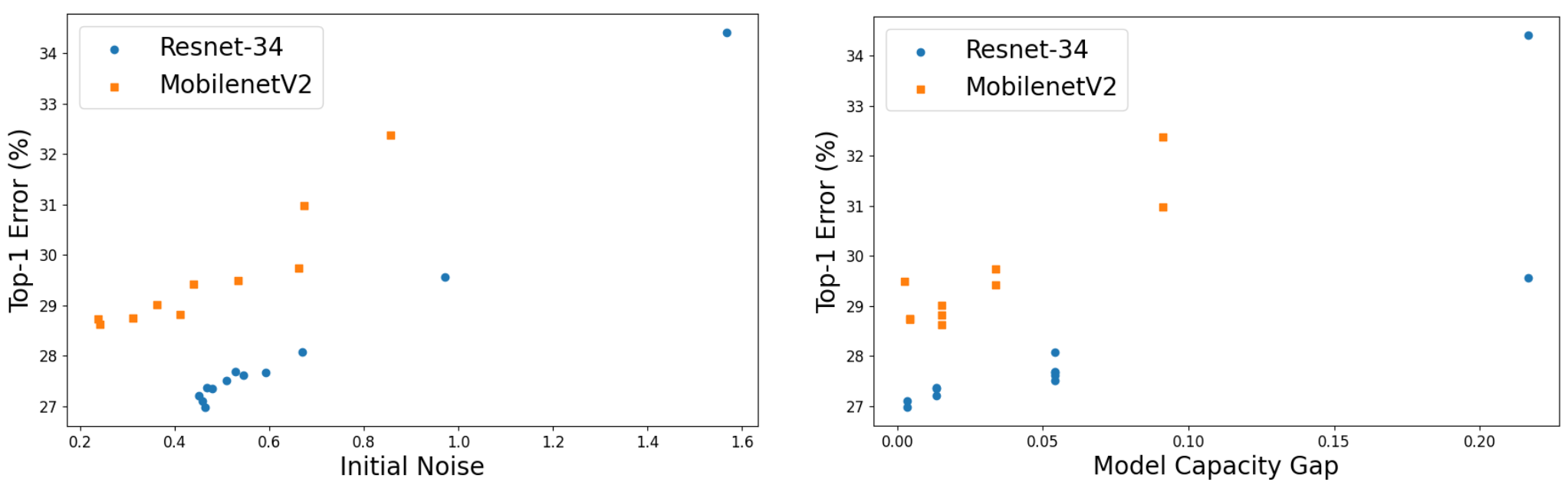}
\end{center}
   \caption{(left) The initial noise (measured by Equation~\ref{eq:output noise}) and the finetuned error (\%) on ImageNet-1k. Each data point represents the removal of one block, followed by 2000 finetuning steps for the pruned model. (right) Model capacity gap (measured by Equation~\ref{eq:g_m}) and finetuned error (\%) on ImageNet-1k. Each data point corresponds to removing one block, followed by 2000 finetuning steps.}
\label{fig:Model capacity gap and finetuned accuracy}
\end{figure*}

Here, $P_n$ is the number of pixels in the feature map $F_{n}$, and MSE denotes the mean squared error, averaged across a batch of images. A higher initial noise $\epsilon_{ini}$ is anticipated to result in a higher finetuned error, as confirmed by Figure ~\ref{fig:Model capacity gap and finetuned accuracy} (left).

\textbf{Model capacity gap $G$:} A mismatch in model capacity between student and teacher models has a detrimental effect on distillation performance~\cite{HyunCho2019OnDistillation,ImanMirzadeh2020ImprovedAssistant}. Therefore, the effectiveness of the finetuning process can be quantified by the model capacity gap  $G_j$ in Eq.~\ref{eq:g_m}, where  $|M|$ and $|\overline{M}_j|$ are the original model size  and the pruned model size  (after the removal of block $B_j$):

\begin{equation} \label{eq:g_m}
G_j= \frac{|M|-|\overline{M}_j|}{|M|}
\end{equation}

\noindent  The impact of $G_j$ (Eq.~\ref{eq:g_m}) on the finetuned error is validated by Figure~\ref{fig:Model capacity gap and finetuned accuracy} (right).

\textbf{Prune method:}  To optimizing the tradeoff between test error and inference latency savings from pruning, $\frac{\text{Error} (\overline{M}_{FT})}{\text{Inference Latency Saving}}$ is used to evaluate the blocks' importance.  The $\text{Error} (\overline{M}_{FT})$ can be approximated by the pruning-induced noise $\epsilon_{ini}$ and the efficacy of the fine-tuning process, measured by the model capacity gap $G$. Inference latency saving can be quantified by the normalized latency time saving $\Delta{T}_j$ defined in Eq.\ref{eq:latency reduction}:

\begin{equation} \label{eq:latency reduction}
\Delta{T}_j=\frac{T-\overline{T}_j}{T}
\end{equation}

\noindent where $T$  and $\overline{T}_j$ are the inference latencies of model $M$ and $\overline{M}_j$ (after the removal of block $B_j$ ), and  can be obtained offline by feeding random noised images with the same resolution as test-time samples into the model. Based on this analysis, we propose the following pruning metric to measure the importance $I_j$ of block $B_j$ in Eq.~\ref{eq:prune metric}:

\begin{equation} \label{eq:prune metric}
I_j =\frac{\epsilon_{ini,j} \times G_j}{\Delta{T}_j}
\end{equation}

\noindent The initial noise $\epsilon_{ini}$ for all $n$ blocks is obtained by processing the same batch of images $n$ times. This method bypasses the need for fine-tuning, requiring only $n$ forward passes.

\begin{algorithm}[t]
    \caption{Test Time Model Pruning and Distillation}
    \label{alg: overview}
    \begin{algorithmic}
       \STATE {\bfseries Input:} the pre-trained model $M$, number of pruned blocks $N_p$, and test time images $x$
       \STATE {\bfseries Output:} model $\overline{M}$ pruned and finetuned from model $M$
        \WHILE{True}
        \IF{Pruning unfinished}
            \STATE Infer $x$ with $M$ if new $x$ arrive
            \STATE Prune $M$ by $N_p$ blocks (Eq.~\ref{eq:prune metric})
        \ELSIF{Distillation unfinished}
            \STATE Infer $x$ with $M$ if new $x$ arrive
            \STATE Finetune pruned model via knowledge distillation
        \ELSE
            \STATE Infer x with $\overline{M}$ if new $x$ arrive
            \STATE Self adapt $\overline{M}$ with $x$ (optional)
        \ENDIF
        \ENDWHILE
         \end{algorithmic}
\end{algorithm}

\begin{table}[t]
\centering
  \setlength{\tabcolsep}{1pt}
  \resizebox{0.9\linewidth}{!}{
\begin{tabular}{@{} *{1}{c} *{1}{c} *{1}{c} *{1}{c} @{}}
\toprule
Layer & $P_n$ (Resolution)
& Loss
& Accuracy\\
\midrule
 Final residual block & 25088 &0.1534 & 70.932  \\
 Average pooling layer & 512 &0.0569 &  70.246  \\
\bottomrule
\end{tabular}
}
\caption{Effect of feature map (pseudo label) resolution. Three blocks are pruned from ResNet-34. The pruned model is finetuned by mimicking different feature maps of the teacher model on ImageNet-1K. The feature map resolutions, finetuning loss, and test accuracy (\%) are reported.}
\label{exp: validation of finetune dataset equation}
\end{table}

\subsection{Efficient Test Time Finetuning}
\label{method:finetuning}

\noindent \textbf{Motivation:} While the fine-tuning process, such as knowledge distillation, effectively reduces the prediction errors of the student model (the pruned model)~\cite{Sanh2019DistilBERTLighter,Lagunas2021BlockTransformers,Hou2020DynaBERT:Depth}, it incurs substantial computational costs and increases the fine-tuning time $T_{finetune}$ due to continuous pseudo-label generation by the large teacher model. A straightforward approach to reducing  $T_{finetune}$ is to generate pseudo-labels once for a small set of test-time samples and store them for subsequent fine-tuning of the pruned model. While this reduces pseudo-label generation costs and $T_{finetune}$, using an excessively small dataset for fine-tuning can lead to overfitting issues.  We define the required fine-tuning dataset size ($|D_R|$) as the minimum size below which significant overfitting is likely. The challenge is how to reduce $|D_R|$ for efficient distillation.

The required fine-tuning dataset size is inversely related to the resolution of pseudo labels. In knowledge distillation, the student model mimics the feature map (pseudo labels) of the teacher model, with each pixel in this feature map imposing a constraint on the student model's weights. The constraint strength is thus measured by the total number of feature map pixels ($P_{T} = |D_R| \times P_n$), where $P_n$ is the feature map resolution. To prevent overfitting of the fine-tuning dataset, the student model $\overline{M}$ requires a large enough $P_{T}$ to apply strong enough constraints. This is confirmed by Table~\ref{exp: validation of finetune dataset equation}, where a model fine-tuned with smaller $P_{T}$ (smaller $P_n$ ) exhibits more overfitting tendency, indicated by lower fine-tuning loss but also lower test accuracy. Consequently, to avoid the overfitting issue, the student model size $|\overline{M}|$ determines the required $P_{T}$ ($|D_R| \times P_n$):

\begin{equation} \label{eq:sample_efficiency p_t}
P_{T} = |D_R| \times P_n \propto \overline{M}
\end{equation}

\noindent Therefore,  $|D_R|$ is positively correlated with $|\overline{M}|$ but negatively correlated with resolution $P_n$ (Eq.~\ref{eq:sample_efficiency}):
\begin{equation} \label{eq:sample_efficiency}
|D_{R}| \propto \frac{|\overline{M}|}{P_n}
\end{equation}

\begin{figure}[t]
  \centering
\includegraphics[width=1.0\linewidth,height=2.0in]{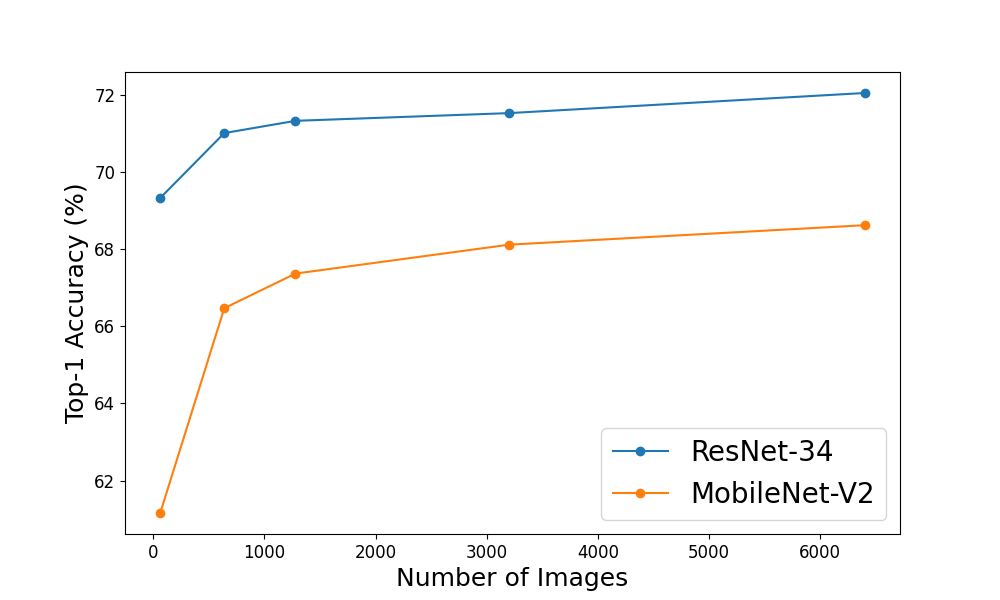}
    \caption{Number of finetuned images and finetuned test accuracy on ImageNet-1k. Three blocks are pruned from ResNet and MobileNet, followed by finetuning for 2000 steps.}
    \label{fig:sample_efficiency}
\end{figure}

\noindent A derivation from these analyses is that a larger $P_n$ reduces $|D_R|$. This is confirmed by Figure~\ref{fig:sample_efficiency}, where a large $P_n$ of 25000 reduces the required dataset size $|D_R|$ to around 3000 images, after which test accuracy starts to saturate. For this reason, we select the large feature map output from the final block as the pseudo labels.  Pseudo-labels for the fine-tuning dataset $D_R$ are generated once and stored in memory. The pruned model is then fine-tuned using the images and their respective pseudo-labels from $D_R$, thereby reducing pseudo labels generation cost and enhancing efficiency.

Using the fine-tuning dataset and corresponding pseudo labels, the fine-tuning process minimizes the mean squared error (loss) between the pseudo labels and the feature map output from the final block of the student (pruned) model.

\section{Experiments}
\textbf{Datasets:} We use a subset of the ImageNet-1k validation set (50,000 images) from ImageNet-1k~\cite{Deng2009Imagenet:Database}, the ImageNet-C~\cite{Hendrycks2019BenchmarkingPerturbations} dataset at the highest noise level, and the ImageNet-R~\cite{Hendrycks2021TheGeneralization} dataset to prune and finetune the model during test time. ImageNet-C consists of 15 corrupted datasets generated by applying 15 different artificial noises (e.g., gaussian noise, snow) to the ImageNet validation set. ImageNet-R contains 16 renditions (e.g., art, cartoons) of 200 ImageNet classes, resulting in 30,000 images.

\textbf{Implementation Details:} ResNet-34~\cite{He2016DeepRecognition}, MobileNetV2~\cite{Sandler2018MobileNetV2:Bottlenecks} and vision transformer ViT-L~\cite{Dosovitskiy2021AnScale} are pruned with 64 images randomly sampled from the validation set of ImageNet-1k/ImageNet-C/ImageNet-R. Only blocks with the same input and output feature map size are pruned.  The pruned models are subsequently fine-tuned via knowledge distillation using 1000 images randomly sampled from the same datasets. During knowledge distillation, all weights except the classifier in the pruned model are optimized for 500 steps using SGD with a batch size of 64  to minimize prune-induced noise. If the fine-tuning dataset size is smaller than the batch size, the batch size equals the dataset size. The fine-tuned models are then evaluated on the remaining validation set images. The initial learning rate is 0.02, and is decayed by 0.1 every 40\% of the steps. Data augmentations are not applied. All experiments are conducted on a single A100 GPU.

\subsection{Main Results}

\textbf{Contenders:} We compare our method with four pruning baselines: (1) A random method that prunes blocks randomly; (2) CURL~\cite{Luo2020NeuralLimited-Data} which removes the block with the least influence on the prediction probability of a model; (3) $l^2$ ratio~\cite{Jastrzebski2018ResidualInference} which prioritizes the removal of blocks where the similarity between input and output feature maps is high; (4) Practise~\cite{Wang2023PracticalSets} which prunes the block with the smallest finetuning loss and largest inference latency time-saving. Models pruned by these baselines are subsequently fine-tuned via knowledge distillation with continuous pseudo labels generation from the teacher model.

 \begin{table}[t]
\centering
  \setlength{\tabcolsep}{1pt}
  \resizebox{\linewidth}{!}{
\begin{tabular}{@{} l l *{3}{c} *{3}{c} *{3}{c} @{}}
\toprule
\multirow{2}{*}{Model} & \multirow{2}{*}{Method}
& \multicolumn{3}{c}{1 block} 
& \multicolumn{3}{c}{3 blocks} 
& \multicolumn{3}{c@{}}{5 blocks}\\
\cmidrule{3-5} \cmidrule(l){6-8} \cmidrule(l){9-11}
& & Accuracy & LS & PF & Accuracy& LS & PF & Accuracy& LS & PF\\
\midrule
\multirow{6}{*}{ResNet-34}
& Random & 71.94 & 3.68 & 138.57 & 69.67  &  12.79  & 151.18 & 57.40 & 24.67 & 167.40 \\
& CURL & 72.64 & 8.48  & 160.69 & 70.62  &  17.60   & 163.74 & 67.36 &  27.13 & 163.70\\
& $l^2$ ratio & 72.11 & 3.75 & 142.99 & 69.71 &  14.49  & 142.68 & 65.80 & 23.60  & 142.61 \\
& Practise & 72.69 & 8.47  & 6046.22 &  71.01 &  22.33   & 6461.67 & 67.11 & 33.07 & 6461.36\\
& Proposed & \textbf{72.81} & \textbf{8.48}  &\textbf{100.00} & \textbf{71.09}  &   \textbf{22.33}  & \textbf{100.00}& \textbf{67.44} & \textbf{33.07}  & \textbf{100.00}\\
\cmidrule{3-11} 
& Source  & 73.31 & 0.00  & 0.00 &  73.31 & 0.00  & 0.00  & 73.31 & 0.00  & 0.00 \\
\midrule
\multirow{6}{*}{MobileNetV2}
& Random & 71.00 & 2.28 & 145.29  & 62.53  & 9.00    & 176.94&57.08  & 24.24   & 169.94\\
& CURL & \textbf{71.08} & 4.64 & 154.08 & 68.79    &  11.43  & 162.49 & 63.90    & 16.49   & 161.94\\
& $l^2$ ratio & 71.04 & 2.28 & 142.96  & \textbf{68.94}  & 9.24 & 152.02  & \textbf{64.05}  &  16.08   & 153.86\\
& Practise & 69.67 & 13.93   & 5020.54 & 65.80  &   22.36   & 5251.94 & 60.36 & 27.61   & 5215.26\\
& Proposed & 69.90 & \textbf{13.93} & \textbf{100.00} & 66.27    & \textbf{22.36} &  \textbf{100.00} & 61.06    &  \textbf{27.69}   & \textbf{100.00}\\
\cmidrule{3-11} 
& Source  & 71.90 & 0.00  & 0.00 &  71.90 & 0.00  & 0.00  & 71.90 & 0.00  & 0.00 \\
\midrule
\multirow{6}{*}{ViT-L/16}
& Random & 78.83 & 3.92 & 102.97 & 77.44  &  12.37 & 103.48 & 57.83 &  20.71  & 103.54\\
& CURL & 79.60 & 3.51 & 118.47 &  68.68 &11.79  & 119.90 & 46.77 & 20.10 &  121.93   \\
& $l^2$ ratio & 79.54 & 3.90 & 103.47 &  79.45 & 12.35 & 103.99 & 77.04 &  20.49  & 104.57\\
& Practise & 79.58 & \textbf{4.14} & 1921.91 &  79.44 & \textbf{12.39} & 2082.06 & \textbf{78.52} &  \textbf{20.73}  & 2262.60\\
& Proposed & \textbf{79.60} &  3.55 & \textbf{100.00} &  \textbf{79.48} & 11.88 & \textbf{100.00} & 78.03 &  20.17  & \textbf{100.00} \\
\cmidrule{3-11} 
& Source  & 79.69 & 0.00  & 0.00 & 79.69  & 0.00 & 0.00 & 79.69 &  0.00  & 0.00 \\
\bottomrule
\end{tabular}
}
\caption{Comparison of test time pruning and finetuning methods on ImageNet-1k. The test accuracy (Accuracy, \%), latency time saving (LS, \%) normalized by the latency of the unpruned model, and pruning and finetuning time (PF,\%) normalized by the time of the proposed method are reported. Source denotes the original pre-trained model.}
\label{exp: main results/accuracy latency tradeoff}
\end{table}

\begin{table}[t]
\centering
  \setlength{\tabcolsep}{12pt}
  \resizebox{0.8\linewidth}{!}{
\begin{tabular}{@{} *{1}{c} *{1}{c} *{1}{c} *{1}{c} @{}}
\toprule
%  Time &  \multicolumn{3}{l}{$t\xrightarrow{\hspace*{4cm}}$}\\
% \hline
Method & Pruning
& Finetuning
& Inference\\
\midrule
Random & \textbf{0.00} & 90.98 &  122.65 \\
CURL & 11.34 & 98.54 & 128.47 \\
 $l^2$ ratio & 0.59 & 85.87 & 116.92 \\
 Practise & 3803.54 & 3888.79 & 3917.00\\
  Proposed & 12.05 & \textbf{48.14} & \textbf{76.34}\\
\bottomrule
\end{tabular}
}
\caption{ The elapsed time (in seconds) at the completion of pruning, fine-tuning, and inference. Three blocks are pruned from ResNet-34. All methods prune and fine-tune the model at test time. }
\label{exp: elapsed time}
\end{table}

Compared to the baselines, our method achieves a similar or better tradeoff between test accuracy and latency time saving while significantly improving pruning and finetuning efficiency. Table \ref{exp: main results/accuracy latency tradeoff} demonstrates that thanks to the proposed pruning method in Equation \ref{eq:prune metric}, our method produces a comparable accuracy and inference latency saving tradeoff with the strongest baseline, practise, for ResNet, MobileNet, and ViT-L across various numbers of pruned blocks. Although some baselines, such as CURL and $l^2$ ratio, achieve around 2-3\% higher accuracy in certain MobileNet experiments, they yield approximately 10\% lower latency saving, providing a less favorable tradeoff compared to our method. Table~\ref{exp: main results/accuracy latency tradeoff} also shows this superior tradeoff is achieved with an averaged 32\% reduction in pruning and finetuning time (PF columns) compared to $l^2$ ratio, the fastest pruning and fine-tuning method. The time saving is much more substantial compared to practise, which has a similar accuracy and latency tradeoff with our method. This should be attributed to our efficient pruning and finetuning approach. The pruning and finetuning time savings are smaller for ViT-L primarily because the time saving from the reduced pseudo-label generation cost is minimal compared to the substantial fine-tuning cost of the pruned ViT-L with over 160 million parameters.

Table~\ref{exp: elapsed time} presents the total elapsed time for pruning, fine-tuning, and inference completed sequentially. While our method achieves comparable pruning efficiency to CURL, it is slower than the $l^2$ ratio method, which requires only one forward pass compared to $n$ forward passes in our method. However, our method requires the least time for fine-tuning the pruned model and results in a model with higher inference speed compared to those pruned by baseline methods. Consequently, our approach completes fine-tuning, and inference in significantly less time than the baselines.

\subsection{Comparison with few/zero shot pruning methods}

\textbf{Contenders:} We compare our method with three few/zero-shot pruning baselines: (1) Practise~\cite{Wang2023PracticalSets}, a few-shot pruning approach that prunes the blocks of the pre-trained model with only a few images sampled from the original training dataset, (2) $l_{2}$-GM~\cite{He2019FilterAcceleration}, a data-free channel pruning approach and (3) Merge~\cite{Kim2020NeuronNeurons}, a data-free approach that merges similar channels.

Our method's performance is not dependent on access to the original training set. As shown in Table \ref{exp: few/zero shot pruning}, Practise~\cite{Wang2023PracticalSets} exhibits declining accuracy with reduced fine-tuning dataset size due to overfitting. Zero-shot approaches ($l_{2}$-GM~\cite{He2019FilterAcceleration} and Merge~\cite{Kim2020NeuronNeurons}) show an even poorer tradeoff between accuracy and latency savings, partly because channel pruning offers a worse tradeoff than block pruning~\cite{Wang2023PracticalSets}. In contrast, our method does not rely on samples from the original training dataset for pruning and fine-tuning, ensuring its performance remains unaffected by access to the training dataset. More results are provided in the supplementary materials.

\subsection{Comparison on the shifted test time distribution}
\label{sec: Test Time distribution shift}

We compare two pruning approaches (train time pruning and test time pruning) when test data distribution differs significantly from train data distribution. The train time pruning approach prunes and fine-tunes ResNet-34 on the training dataset (ImageNet-1k). The test time pruning approach prunes and fine-tunes the same model on the test data (ImageNet-C or ImageNet-R). All fine-tuned models are then evaluated for accuracy on the test data (ImageNet-C or ImageNet-R).

\begin{table}[t]
\vspace{0pt}
\centering
\setlength{\tabcolsep}{1.0pt}
\resizebox{\linewidth}{!}{
\begin{tabular}{@{} l l *{2}{c} *{2}{c} *{2}{c} @{}}
\toprule
\multirow{2}{*}{Model} & \multirow{2}{*}{Method}
& \multicolumn{2}{c}{20 images} 
& \multicolumn{2}{c}{128 images} 
& \multicolumn{2}{c@{}}{1000 images}\\
\cmidrule{3-4} \cmidrule(l){5-6} \cmidrule(l){7-8}
& & Accuracy & LS & Accuracy& LS & Accuracy& LS\\
\midrule
\multirow{4}{*}{ResNet-34}
& Practise & 64.99 & 22.33 & 69.21 & 22.33 &  70.64 & 22.33 \\
& $l_{2}$-GM  & 29.74 & 8.01 & 29.74 &  8.01  & 29.74 & 8.01  \\
& Merge & 32.19 & 0.00 & 32.19 & 0.00   & 32.19 & 0.00 \\
& Proposed & \textbf{71.20} & \textbf{22.33} &  \textbf{71.20} & \textbf{22.33}  &  \textbf{71.20} & \textbf{22.33} \\
\midrule
\multirow{2}{*}{MobileNetV2}
& Practise & 53.87 & 22.36 & 62.17 &  22.36  & 65.46 & 22.36 \\
% & $l_{2}$-GM   \\
% & Merge &  &  &  &    &  & \\
& Proposed & \textbf{66.43} & \textbf{22.36}  & \textbf{66.43}  & \textbf{22.36}    & \textbf{66.43}  & \textbf{22.36}  \\
\bottomrule
\end{tabular}
}
\caption{Practise prunes and finetunes the pre-trained model using 20, 128, and 1000 images from the original training dataset, while the proposed method uses 1000 test time samples. Top-1 test accuracy (\%) on ImageNet-1k and inference latency time saving (LS, \%) normalized by the latency of the pre-trained model are reported.}
\label{exp: few/zero shot pruning}
\end{table}

\begin{table}[t]
\vspace{0pt}
\centering
\setlength{\tabcolsep}{1pt}
\resizebox{\linewidth}{!}{
\begin{tabular}{@{} l l *{2}{c} *{2}{c} *{2}{c} @{}}
\toprule
\multirow{2}{*}{Model} & \multirow{2}{*}{Prune \& Finetune (PF)}
& \multicolumn{2}{c}{1 block} 
& \multicolumn{2}{c}{2 blocks} 
& \multicolumn{2}{c@{}}{3 blocks}\\
\cmidrule{3-4} \cmidrule(l){5-6} \cmidrule(l){7-8}
& & Accuracy & LS & Accuracy & LS& Accuracy & LS\\
\midrule
\multirow{6}{*}{ImageNet-C}  & Train-Time PF  & 17.21 & 8.69 & 16.73 & 17.30 &  15.58 & 22.78\\
& Test-Time PF & \textbf{19.84} & \textbf{8.80} & \textbf{19.32} & \textbf{17.53} &  \textbf{18.68} & \textbf{23.00}  \\
\cdashline{3-8}
 & Train-Time  PF+TENT& 10.64 & -70.50 & 10.58 & -54.84 &  9.76 &  -50.30 \\
 & Test-Time  PF+TENT& \textbf{13.86} & -69.05 & \textbf{13.98} & -52.69 & 
\textbf{12.81} &  -48.50\\
\cdashline{3-8}
  & Train-Time  PF+EATA& \textbf{25.01} & -70.07 & \textbf{23.87} & -50.70 & 21.80  &  -46.43\\
& Test-Time  PF+EATA& 24.21 & -64.11 & 23.34 & -45.92 &  \textbf{21.90}  &  -42.45 \\
\midrule
\multirow{6}{*}{ImageNet-R}  & Train-Time PF  & 36.14 & 8.81 & 35.88 & 17.29 &  34.97 & 22.84\\
& Test-Time PF   & \textbf{37.59} & \textbf{8.87} & \textbf{36.71} & \textbf{17.41} &  \textbf{35.96} &  \textbf{22.89}  \\
\cdashline{3-8}
 & Train-Time PF+TENT& 36.77 & -74.19 & 36.62 & -54.84 &  35.43 & -56.21 \\
& Test-Time PF+TENT& \textbf{38.17}& \textbf{-70.97} & \textbf{38.24} & \textbf{-53.13} &  \textbf{37.35} &   \textbf{-48.39} \\
 \cdashline{3-8}
  & Train-Time  PF+EATA& 36.59  & -83.87& 36.31 &-64.52 & 35.47  & -53.55 \\
& Test-Time  PF+EATA& \textbf{37.18} & \textbf{-80.64} & \textbf{37.62} & \textbf{-58.06} &  \textbf{37.01}  &  \textbf{-52.50}\\
\bottomrule
\end{tabular}}
\caption{Performance comparison of pruning and fine-tuning using training data vs. test data. Test accuracies (\%) on ImageNet-R and ImageNet-C (averaged on 15 corrupted datasets), and inference latency saving (LS, \%) normalized by the latency of the pre-trained model are reported.}
\label{exp: main results/test time finetuning}
\end{table}

\begin{figure}[t]
\begin{center}
\includegraphics[width=1.0\linewidth,height=2.0in]{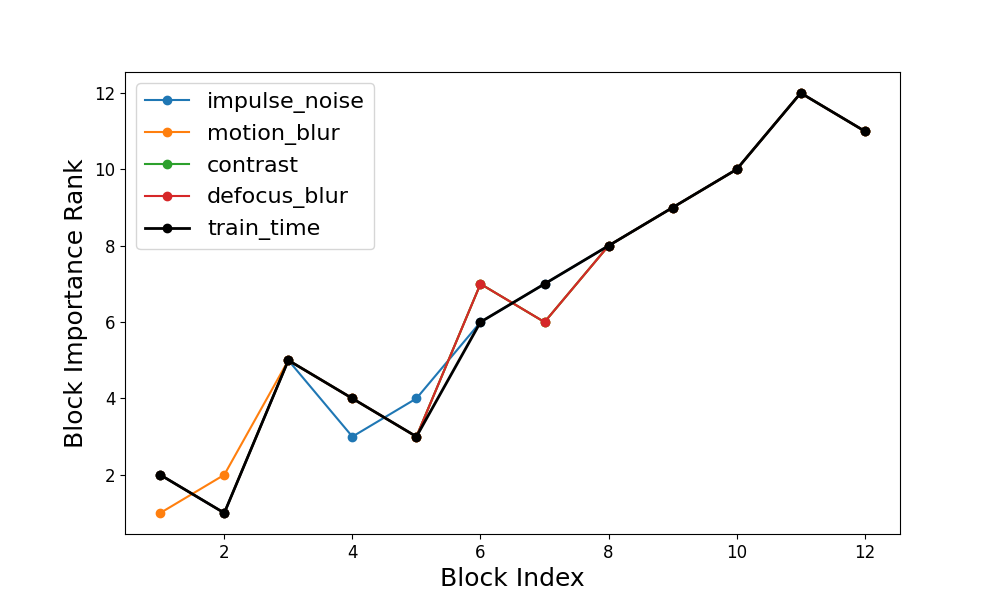}
\end{center}
   \caption{The impact of test time distribution shift on pruning decisions. The importance of 12 removable blocks in ResNet-34, pre-trained on ImageNet-1K, is ranked by the train time dataset (ImageNet-1K) and four different test time datasets in ImageNet-C.}
\label{fig: block importance rank}
\end{figure}

When the test-time distribution shifts from the training-time distribution, pruning decisions based on the training dataset may misalign with those based on test data. To demonstrate this, we compare two pruning strategies for the pre-trained ResNet-34: one based on the original training dataset and the other based on test samples from ImageNet-C (corrupted ImageNet datasets with 15 different noise). As shown in Figure~\ref{fig: block importance rank}, pruning decisions in 4 out of these 15 test-time distribution shifts differ from those made using the training dataset. This finding highlights the necessity of test-time pruning.

In addition to different pruning decisions, finetuning the model via knowledge distillation on the test time samples also achieve higher accuracy over finetuning on train samples. The knowledge transfer from the teacher model to the student model (pruned model) is influenced by the input samples. If the samples come from the training data, the student model learns knowledge specific to the training domain, which may not generalize well to a different test domain. Conversely, when input samples are from the test domain, the student model acquires domain-specific knowledge from the teacher model, enhancing its performance in that domain.  This is confirmed by the comparison between  ``Train-Time PF" and ``Test-Time PF" in Table~\ref{exp: main results/test time finetuning} where test time pruning and finetuning (PF) offers an average 1.93\% higher accuracy than train time counterparts. This finding further highlights the necessity of test-time pruning and finetuning. Additional results are provided in the supplementary materials.

To investigate the impact of test time adaptation techniques such as TENT~\cite{Wang2021Tent:Minimization} and EATA~\cite{Niu2022EfficientForgetting}, we self-train the distilled (finetuned) model on the test data, and  denote these methods as ``Train-Time PF + TENT/EATA" and ``Test-Time PF + TENT/EATA" in Table~\ref{exp: main results/test time finetuning}. We have two key observations. First, while TENT and EATA enhance the test accuracy of the fine-tuned model, they impose a significant computational burden at test time, reducing inference speed by approximately 50\% compared to the unpruned pre-trained model. Second, Test-Time PF + TENT/EATA achieves accuracy comparable to or better than Train-Time PF + TENT/EATA. These findings highlight the importance of test-time pruning and fine-tuning.

\subsection{Ablation study}

\begin{table}[t]
\centering
\setlength{\tabcolsep}{0.5pt}
\resizebox{1.0\linewidth}{!}{
\begin{tabular}{@{} l l *{2}{c} *{2}{c} *{2}{c} @{}}
\toprule
\multirow{2}{*}{Model} & \multirow{2}{*}{Method}
& \multicolumn{3}{c}{3 blocks} 
& \multicolumn{3}{c@{}}{5 blocks}\\
\cmidrule{3-5} \cmidrule(l){6-8}
& & PT & Accuracy & LS & PT & Accuracy & LS\\
\midrule
\multirow{5}{*}{ResNet-34}
& Random & \textbf{0.00} & 68.92 & 13.95  &  \textbf{0.00}  & 56.20 &  20.06 \\
& CURL & 99.13 & 70.69 & 18.30  &  101.09  & 67.08 &  25.83 \\
& $l^2$ ratio & 5.34 & 70.00  &  14.49 &  5.15  & 65.96 & 23.60 \\

& Practise & 34303 & 71.05 & 22.33  &  35308  & 67.29 & 33.07 \\
& Proposed & 100.00 & \textbf{71.12} & \textbf{22.33}  & 100.00   & \textbf{67.54} & \textbf{33.07} \\
\midrule
\multirow{5}{*}{MobileNetV2}
& Random & \textbf{0.00} & 61.11 & 8.75  & \textbf{0.00}   & 58.00 &  22.61\\
& CURL & 98.65 &  68.86 & 11.43 &  107.21  & \textbf{64.39} &  16.08\\
& $l^2$ ratio & 8.61 & \textbf{69.03} &  10.14 &  7.59  & 64.36 & 16.08 \\

& Practise & 46808 & 65.98 & 22.38  &  41985  & 60.21 & 27.68 \\
& Proposed & 100.00 & 66.35 & \textbf{25.41}  & 100.00   & 61.70 & \textbf{27.69} \\
\bottomrule
\end{tabular}
}
\caption{Comparison of pruning methods with the same finetuning approach. Top-1 test accuracy (\%) on ImageNet-1k, inference latency saving (LS, \%), and pruning time (PT, \%) normalized by that of the proposed method are reported.}
\label{ablation: Effect of pruning}
\end{table}

\textbf{Effect of Test Time Pruning:} We compare our pruning approach with prior pruning methods using the same finetuning approach. Table \ref{ablation: Effect of pruning} shows that our pruning method offers a similar or superior tradeoff between test accuracy and latency time saving compared to existing pruning methods. Regarding pruning efficiency, the proposed method is similar to CURL\cite{Luo2020NeuralLimited-Data}, both of which are slower than random and $l^2$ ratio but much faster than practise~\cite{Wang2023PracticalSets}. The main computational cost for the proposed pruning approach is the determination of the prune-induced noise (loss) for each block, requiring $n$ forward processes for $n$ removable blocks. By contrast, the $l^2$ ratio only needs 1 forward process, while random does not need any network computation.

\textbf{Effect of Test Time Finetuning:} We compare our fine-tuning approach to previous methods using the same pruning strategy. Table \ref{ablation: Effect of Finetuning} demonstrates that our fine-tuning method achieves similar or higher accuracy compared to the baselines while being approximately 1.6 times faster. This efficiency stems from our method's use of a small set of stored pseudo-labels and image data, eliminating the need for continual pseudo-label generation by the large teacher network. In contrast, MiR~\cite{Wang2022CompressingReplacing} requires querying the teacher network for every input image, leading to slower fine-tuning speed. Entropy~\cite{Xu2023Computation-EfficientMixup,Wang2020NeuralModel} only queries the teacher model for inputs the student model finds challenging, yet this does not result in significant speedup due to the additional cost of sample filtering. Moreover, the model finetuned by Entropy may be overfitted to these challenging samples.

\begin{figure}[t]
  \centering
\includegraphics[width=1.0\linewidth,height=2.0in]{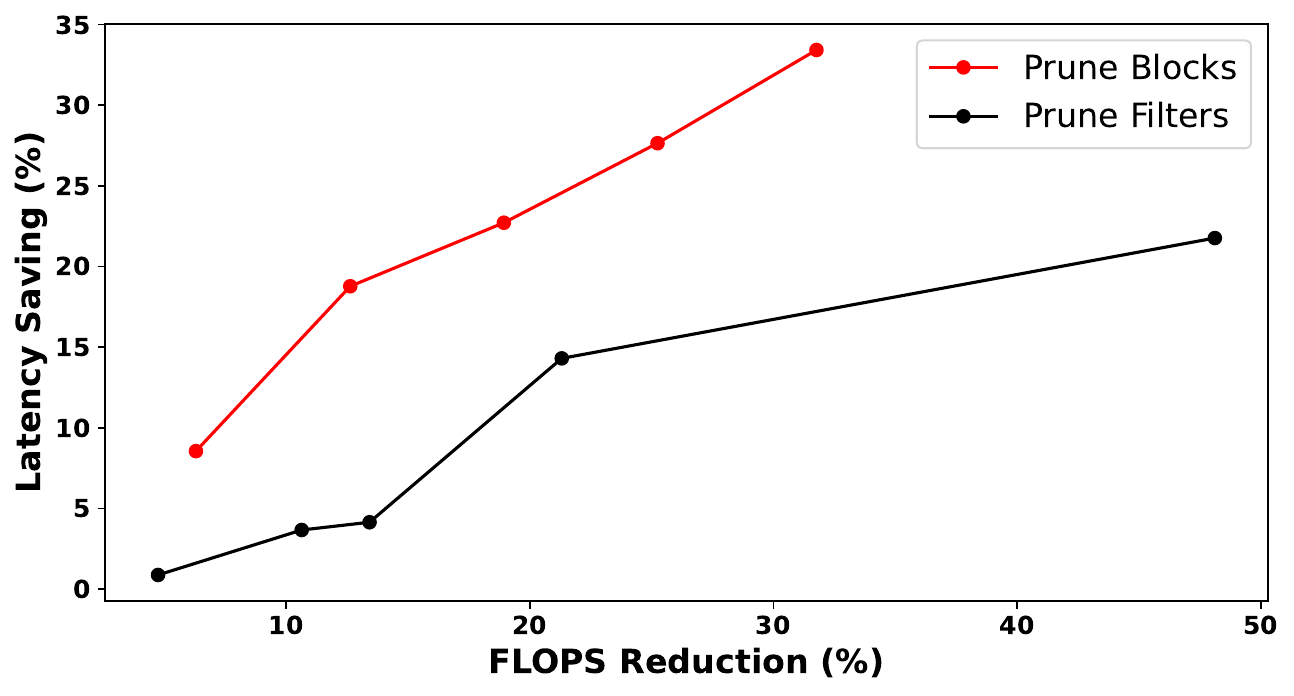}
    \caption{The relationship between latency saving (\%) and FLOPs reduction (\%). Filters or blocks in ResNet34 are pruned using $l_{2}$-GM and the proposed method. }
    \label{fig:filters_or_blocks}
\end{figure}

\textbf{Analysis of Pruning Granularity:} Figure~\ref{fig:filters_or_blocks} demonstrates that pruning blocks results in greater latency savings than pruning filters for a given FLOPs (number of floating point operations) reduction. Consequently, block pruning retains more parameters from the original model for the same latency reduction, potentially enhancing accuracy~\cite{Wang2023PracticalSets}. Thus, block pruning is employed in this study.

\begin{table}[t]
\centering
  \setlength{\tabcolsep}{0.5pt}
\resizebox{1.0\linewidth}{!}{
\begin{tabular}{@{} l l *{2}{c} *{2}{c} *{2}{c} @{}}
\toprule
\multirow{2}{*}{Model} & \multirow{2}{*}{Method}
& \multicolumn{2}{c}{1 block} 
& \multicolumn{2}{c}{3 blocks} 
& \multicolumn{2}{c@{}}{5 blocks}\\
\cmidrule{3-4} \cmidrule(l){5-6} \cmidrule(l){7-8}
& & Accuracy & FT & Accuracy & FT& Accuracy & FT\\
\midrule
\multirow{3}{*}{ResNet-34}
& MiR & 72.78 & 169.59  &  70.84 &   190.29 & 66.87 &  184.42 \\
& Entropy & 72.36 & 167.52 & 70.25  &  175.83  & 65.58 & 191.85 \\
& Proposed &\textbf{72.84}  & \textbf{100.00} &  \textbf{71.15} &  \textbf{100.00}  & \textbf{67.25} & \textbf{100.00} \\
\midrule
\multirow{3}{*}{MobileNetV2}
& MiR & 69.64 & 162.12 &  66.21 & 166.20   & 61.09 & 175.37 \\
& Entropy & 69.45 & 163.12 & 65.59  &  171.61  & 59.63 & 173.29 \\
& Proposed & \textbf{69.75} & \textbf{100.00} & \textbf{66.35}  & \textbf{100.00}  & \textbf{61.15}  & \textbf{100.00} \\
\bottomrule
\end{tabular}
}
\caption{Performance comparison on finetuning methods. We compare MiR~\cite{Wang2022CompressingReplacing}, Entropy~\cite{Xu2023Computation-EfficientMixup}, and our method in cases where 1 block, 3 blocks, and 5 blocks are dropped. We employ our pruning strategy for all baselines. Top-1 test accuracy (\%) on ImageNet-1k, and finetuning time (FT, \%) normalized by that of the proposed method are reported.  }
\label{ablation: Effect of Finetuning}
\end{table}

\section{Conclusions}
We propose a test time pruning and fine-tuning framework. To enhance pruning efficiency, we identify two proxy variables (prune-induced noise and model capacity gap) to approximate fine-tuned accuracy. These proxies are used to evaluate blocks. To improve finetuning efficiency, we use high dimensional pseudo labels to reduce the required finetuning dataset size. Experiments demonstrate that compared to baselines, our approach offers a similar or better tradeoff between test accuracy and inference latency while saving around 32\% of pruning and finetuning time. 

\textbf{Limitations:}  The pruning method requires multiple forward passes to obtain initial noise, and the fine-tuning method demands several hundred fine-tuning steps. There is potential for efficiency improvements.

\bibliography{CameraReady/main}

\end{document}

\maketitle

\begin{table*}[ht!]
\centering
  \setlength{\tabcolsep}{1.0pt}
  \resizebox{1.0\linewidth}{!}{
\begin{tabular}{@{} l l *{2}{c} *{2}{c} *{2}{c} *{2}{c} *{2}{c} *{2}{c}@{}}
\toprule
\multirow{2}{*}{Model} & \multirow{2}{*}{Method}
& \multicolumn{2}{c}{20 images} 
& \multicolumn{2}{c}{40 images} 
& \multicolumn{2}{c@{}}{80 images}
& \multicolumn{2}{c@{}}{128 images}
& \multicolumn{2}{c@{}}{512 images}
& \multicolumn{2}{c@{}}{1000 images}\\
\cmidrule{3-4} \cmidrule(l){5-6} \cmidrule(l){7-8} \cmidrule(l){9-10} \cmidrule(l){11-12} \cmidrule(l){13-14}
& & Accuracy & LS & Accuracy& LS & Accuracy& LS & Accuracy& LS & Accuracy& LS & Accuracy& LS \\
\midrule
\multirow{4}{*}{ResNet-34}
& Practise & 64.99 & 22.33 & 66.94 & 22.33 &  67.792 & 22.33 & 69.21 & 22.33 &70.37 & 22.33 &70.64 & 22.33\\
& $l_{2}$-GM  & 29.74 & 8.01 & 29.74 &  8.01  & 29.74 & 8.01  & 29.74 &  8.01  & 29.74 &  8.01  & 29.74 &  8.01  \\
& Merge & 32.19 & 0.00 & 32.19 & 0.00   & 32.19 & 0.00 & 32.19 & 0.00 & 32.19 & 0.00 & 32.19 & 0.00\\
& Proposed & \textbf{71.20} & \textbf{22.33} &  \textbf{71.20} & \textbf{22.33}  &  \textbf{71.20} & \textbf{22.33} &  \textbf{71.20} & \textbf{22.33}  &  \textbf{71.20} & \textbf{22.33}  &  \textbf{71.20} & \textbf{22.33}  \\
\midrule
\multirow{2}{*}{MobileNetV2}
& Practise & 53.87 & 22.36 & 58.11 &  22.36  & 61.31 & 22.36 & 62.17 &  22.36  & 65.00 & 22.36 & 65.46 & 22.36 \\
% & $l_{2}$-GM   \\
% & Merge &  &  &  &    &  & \\
& Proposed & \textbf{66.43} & \textbf{22.36}  & \textbf{66.43}  & \textbf{22.36}    & \textbf{66.43}  & \textbf{22.36}  & \textbf{66.43}  & \textbf{22.36}  & \textbf{66.43}  & \textbf{22.36}  & \textbf{66.43}  & \textbf{22.36}   \\
\bottomrule
\end{tabular}
}
\caption{Practise~\cite{Wang2023PracticalSets} prunes and finetunes the pretrained model using 20 images, 40 images, 80 images, 128 images,  512 images and 1000 images sampled from the original train dataset, while the proposed method is pruned and finetuned on 1000 test time samples. Top-1 test accuracy (\%) on ImageNet-1k and inference latency time saving (LS, \%) are reported.}
\label{appendix: few/zero shot pruning}
\end{table*}

\begin{table*}[t]
\begin{center}
\scalebox{1.0}{
 \setlength{\tabcolsep}{1.0mm}{
\begin{tabular}{lccccccccccccccc}
\hline
& \rotatebox{90}{brightness} & \rotatebox{90}{contrast} & \rotatebox{90}{defocus blur} & \rotatebox{90}{elastic transform} & \rotatebox{90}{fog} & \rotatebox{90}{frost} & \rotatebox{90}{gaussian noise} & \rotatebox{90}{glass blur} & \rotatebox{90}{impulse noise} & \rotatebox{90}{jpeg compression} & \rotatebox{90}{motion blur} & \rotatebox{90}{pixelate} & \rotatebox{90}{shot noise} & \rotatebox{90}{snow} & \rotatebox{90}{zoom blur}\\
\hline
\multicolumn{16}{c}{1 block}\\
\hline
Train-Time PF & 51.32 & 3.45 & 12.85 & 16.67 & 18.09 & 17.40 & 2.98 & 9.38 & 2.93 & 38.10 & 13.25 & 34.12 & 4.13 & 13.83 & 19.64\\
Test-Time PF & \textbf{54.72} & \textbf{4.10} & \textbf{15.18} & \textbf{20.03} & \textbf{23.97} & \textbf{21.20} & \textbf{5.12} & \textbf{9.92} & \textbf{6.11} & \textbf{41.69} & \textbf{13.86} & \textbf{36.99} & \textbf{6.57} & \textbf{16.44} & \textbf{21.75}\\

Train-Time PF+TENT  & 57.79 & 0.20 & 2.07 & 2.67 & 1.39 & 5.42 & 0.24 & 1.17 & 0.26 & 37.02 & 2.87 & 39.67 & 0.37 & 1.90 & 6.48 \\
Test-Time PF+TENT  & \textbf{58.48} & \textbf{1.00} & \textbf{8.30} & \textbf{5.10} & \textbf{8.02} & \textbf{9.49} & \textbf{0.52} & \textbf{5.49} & \textbf{0.80} & \textbf{42.82} & \textbf{4.21} & \textbf{42.73} & \textbf{0.73} & \textbf{4.79} & \textbf{15.37}\\
\hline
\multicolumn{16}{c}{2 blocks}\\
\hline
Train-Time PF  &49.68 & 1.69 & 11.00 & 18.65 & 17.34 & 15.966 & 2.68 & 10.87 & 2.83 & 39.41 & 13.44 & 30.74 & 3.99 & 12.84 & 19.74\\
Test-Time PF & \textbf{53.80} & \textbf{3.33} & \textbf{15.01} & \textbf{19.29} &\textbf{22.75} & \textbf{20.79} & \textbf{5.12} & \textbf{9.70} & \textbf{5.97} & \textbf{41.45} & \textbf{13.59} & \textbf{35.94} & \textbf{6.37} & \textbf{15.68} & \textbf{21.07}\\
Train-Time PF+TENT  & 56.54 & 0.18 & 1.90 & 3.07 & 1.28 & 3.82 & 0.21 & 1.08 & 0.14 & 38.99 & 3.26 & 37.71 & 0.31 & 1.772 & 8.42\\
Test-Time PF+TENT  & \textbf{57.84} & \textbf{0.63} & \textbf{8.18} & \textbf{6.32} & \textbf{6.79} & \textbf{9.51} & \textbf{0.55} & \textbf{5.39} & \textbf{0.72} & \textbf{42.84} & \textbf{5.47} & \textbf{41.24} & \textbf{0.76} & \textbf{5.93} & \textbf{17.44}\\
\hline
\multicolumn{16}{c}{3 blocks}\\
\hline
Train Time  &47.35 & 1.18 & 10.54 & 18.89 & 15.04 & 14.45 & 1.96 & 9.70 & 2.41 & 36.33 & 12.07 & 30.78 & 3.21 & 11.48 & 18.38\\
Test-Time PF &\textbf{51.44} & \textbf{3.77} & \textbf{14.59} & \textbf{18.85} & \textbf{22.36} & \textbf{19.84} & \textbf{4.88} & \textbf{9.27} & \textbf{5.69} & \textbf{40.34} & \textbf{13.30} & \textbf{34.76} & \textbf{6.05} & \textbf{14.88} & \textbf{20.20}\\
Train Time+TENT  & 53.92 & 0.14 & 1.81 & 3.70 & 0.98 & 2.15 & 0.19 & 1.97 & 0.22 & 35.68 & 2.62 & 36.36 & 0.28 & 1.77 & 4.46\\
Test-Time PF+TENT  & \textbf{56.22} & \textbf{0.65} & \textbf{8.31} & \textbf{4.97} & \textbf{5.63} & \textbf{7.71} & \textbf{0.55} & \textbf{5.21} & \textbf{0.71} & \textbf{41.45} & \textbf{4.45} & \textbf{38.70} & \textbf{0.75} & \textbf{4.35} & \textbf{12.40}\\
\hline
\end{tabular}}}
\end{center}
\caption{ Performance comparison of train-time pruning an fine-tuning (PF) and test-time PF. Train-time PF prunes and fine-tunes ResNet-34 on the training dataset (ImageNet-1k), while test-time pruning PF prunes and fine-tunes the same model on the test data (15 corrupted datasets in ImageNet-C). Test accuracies (\%)  on 15 corrupted datasets of ImageNet-C are reported. }
\label{table:more results on the comparison of train-time and test-time pruning}
\end{table*}

\section{Comparison with few/zero shot pruning}
Our method's performance is independent of access to the original training dataset. As illustrated in Table~\ref{appendix: few/zero shot pruning}, Practise~\cite{Wang2023PracticalSets} demonstrates reduced accuracy with smaller fine-tuning datasets due to overfitting. Zero-shot methods such as  ($l_{2}$-GM~\cite{He2019FilterAcceleration} and Merge~\cite{Kim2020NeuronNeurons}) exhibit an even less favorable tradeoff between accuracy and latency savings~\cite{Wang2023PracticalSets}. In contrast, our method maintains its performance without relying on samples from the original training dataset for pruning and fine-tuning. 

\section{Comparison on the shifted data distribution}

In this section, we compare Test-time Pruning and Fine-tuning (PF) with Train-time PF, as presented in Table~\ref{table:more results on the comparison of train-time and test-time pruning} for the 15 corrupted datasets in ImageNet-C. The results indicate that Test-time PF consistently outperforms Train-time PF in accuracy across all 15 distribution shifts without test-time adaptation. Furthermore, when the pruned and fine-tuned model is adapted using test time techniques such as TENT~\cite{Wang2021Tent:Minimization}, Test-time PF still outperforms Train-time PF. This underscores the importance of Test-time Pruning and Fine-tuning.

\bibliography{CameraReady/main}